%% file: gcrl_workshop.tex
\documentclass{article}

\input{math_commands.tex}

\usepackage[final]{neurips_2023}

\usepackage[utf8]{inputenc} 
\usepackage[T1]{fontenc}    
\usepackage{hyperref}       
\usepackage{url}            
\usepackage{booktabs}       
\usepackage{amsfonts}       
\usepackage{nicefrac}       
\usepackage{microtype}      
\usepackage{xcolor}         
\usepackage{graphicx}
\usepackage{subcaption}
\usepackage{multirow}
\usepackage{amsmath}

\definecolor{orangefield}{rgb}{1.0,0.5,0}
\definecolor{cyanfield}{rgb}{0,0.5,0.5}
\definecolor{yellowfield}{rgb}{1.0,1.0,0.0}
\definecolor{violetfield}{rgb}{1.0,0.0,1.0}

\definecolor{greenfield}{rgb}{0.0,0.39,0.15}

\definecolor{orangegoal}{rgb}{0.75,0.37,0}
\definecolor{cyangoal}{rgb}{0.0,0.65,0.65}
\definecolor{yellowgoal}{rgb}{0.71,0.71,0}
\definecolor{violetgoal}{rgb}{0.7,0.0,0.7}

\title{Backward Learning for Goal-Conditioned Policies}

\author{Marc Höftmann,$\:$ Jan Robine,$\:$  Stefan Harmeling  \\
	Department of Computer Science, Technical University of Dortmund, Germany}

\begin{document}

\maketitle
 
\begin{abstract}
    Can we learn policies in reinforcement learning without rewards?  Can we learn a policy just by trying to reach a goal state?  We answer these questions positively by proposing a multi-step procedure that first learns a world model that goes backward in time, secondly generates goal-reaching backward trajectories, thirdly improves those sequences using shortest path finding algorithms, and finally trains a neural network policy by imitation learning.  We evaluate our method on a deterministic maze environment where the observations are $64\times 64$ pixel bird's eye images and can show that it consistently reaches several goals. Our code is available at {\tiny \url{https://github.com/Hauf3n/Backward-Learning-for-Goal-Conditioned-Policies}}.

\end{abstract}
\begin{center}
\end{center}

\section{Motivation}
Recently, generative auto-regressive models have shown great prospect with the developments around GPT-3 \citep{brown2020language} where predictive coding \citep{huang2011predictive} seems a promising avenue for further advancements in machine learning.  Model-based Reinforcement learning (MBRL) applies the same principle already for decades \citep{schmidhuber1991possibility} in order to learn the dynamics of an environment which is then exploited to maximize the return by learning a policy. These models can have different advantages to facilitate policy optimization, e.g., simulate future outcomes \citep{kaiser2019model,hafner2023mastering,robine2023transformer}, conduct online planning \citep{schrittwieser2020mastering} or provide the model's state memory as an additional policy input \citep{ha2018world}.

But sometimes no reward function is given and instead some goal states have to be reached. An important variant of reinforcement learning (RL) is \emph{goal-conditioned} RL, where a policy tries to arrive at a state configuration. Typical modern approaches \citep{andrychowicz2017hindsight,hafner2022deep,pateria2021hierarchical,ecoffet2021first,li2022hierarchical,gupta2019relay} start at a certain state and then try to reach a selected goal.  We call this paradigm \emph{hit-and-miss}, since the trajectories generated by these approaches within a reasonable amount of steps are not guaranteed to contain a desired goal.

\textbf{Outline.} In our work, we overcome this problem with the help of world models by ensuring that every produced trajectory \emph{does} reach the goal state.  We do this by starting at the goal and then going backward in time.  To achieve this, we collect random forward trajectories on which we train a backwards world model that takes a state-action pair $(s_t,a_{t-1})$ as input and outputs the previous state, e.g., $f(s_t,a_{t-1}) = s_{t-1}$.

Once the backwards world model is sufficiently trained, we use it to learn policies.  In goal-conditioned RL the assumption is that we are given some goal state $s^*$.  In real world problems this goal might be problem specific or be determined by intrinsic or extrinsic reward, game score or by human preference. After the goal selection our backwards world model generates rollouts, reversed in time, that ultimately look like typical forward sequences $(s_0,a_0,\dots,s_{t-1},a_{t-1},s_t,a_t,s_{t+1},a_{t+1},\dots,s^*)$ but without rewards.

In principle these forward trajectories seem ideal to train a policy by imitation learning, since each sequence always reaches the goal state.  However, the simulations should not be carelessly used for optimization due to potential loops or poor action selection \citep[investigated by][]{wu2019imitation,wang2021learning}. Therefore, further tricks are needed to exclude some of the state-actions pairs, which we analyze and determine by graph search over simulated trajectories.
\\

\section{Prior Work}
\textbf{Offline RL.} \citet{chebotar2021actionable} learn a goal-conditioned Q-function on offline data where they use sub-sequences and relabeling techniques to generate numerous of goal trajectories to learn from and failure actions to avoid. \citet{kidambi2020morel} use offline MBRL to learn a pessimistic estimate of the underlying MDP and optimize a policy on it while minimizing the model bias.    

\textbf{Backward world models.} Currently, world models that unroll simulated trajectories into the past (backward models) represent a small part in the model-based reinforcement learning literature. It starts with models that improve sample efficiency \citep{goyal2018recall} or use improved strategies for returning to high value states \citep{edwards2018forward}.
\cite{lai2020bidirectional} proposes bi-directional world models which generate short forward and backward trajectories to be more robust against prediction error of forward models that unroll for too many timesteps.
\cite{wang2021offline} learn a reverse policy and reverse dynamics model for offline reinforcement learning. They generate backward simulations which lead to novel sub-goal reaching trajectories with respect to the existing offline dataset.
\cite{chelu2020forethought} address the RL credit assignment problem and examine the benefits of planning with a backward model in relationship to forward models, where they observe complementary gains using backward models.
\cite{pan2022backward} combine imitation learning with backward world models where backward simulations are treated as sub-optimal expert demonstrations to improve the expected return of a forward policy. In this work, we utilize our discrete backward model in a way which allows it to create demonstrations that can be used without considering the expected return.

\section{Method}
As commonly done in model-based RL, we project the raw state observations into
a latent space. In order to also utilize return strategies from Go-Explore methods \citep{ecoffet2021first, hoftmann2023time} we
learn an encoder $\Phi_\theta$ that generates a \emph{discrete} latent
representation \citep[following][]{hafner2020mastering} jointly with a \emph{latent} world model $\Psi_\theta$. In short, we denote them as
\begin{align}
    \Phi_\theta(s_t) & = Z_t & \Psi_\theta(z_t,a_{t-1}) & = \hat{Z}_{t-1}.
\end{align}
where $Z_t,\hat{Z}_{t-1}$ are distributions over the latent space and discrete samples $z_t$ can be obtained via sampling, i.e., $z_t \sim Z_t$.
In addition, we define the latent space decoder as $s_t \sim p_\theta(s_t|z_t)$.

\textbf{Representation Model.} The representation model is implemented by a Convolutional Neural Network \citep[CNN,][]{lecun1989backpropagation} with an Encoder-Decoder architecture \citep{ballard1987modular} that uses the categorical latent representation from \cite{hafner2020mastering} with straight-through gradients. In our case, the encoder's latent representation $Z \in \mathbb{R}^{g \times c}$ is stochastic and uses $g = 16$ categoricals with $c = 16$ classes each, where we  obtain a discrete representation $z \sim Z$ by sampling from $g$ softmax distributions. The representation loss can be written as

\begin{equation}
	\mathcal{L}^{\text{Repr}}_\theta = \mathbb{E}_t\left[ \underbrace{- \log p_\theta(s_t|z_t)}_\text{reconstruction loss} + \underbrace{\alpha \cdot \max\left(\frac{1}{g c}\sum_{i=1}^{g} \sum_{j=1}^{c} Z_{t_{i,j}} \log(Z_{t_{i,j}}),\:0.05\right)}_\text{entropy loss} \right]\text{with} \:\: z_t \sim Z_t,
\end{equation}
and consists of two objectives. The first one is the classic reconstruction loss to minimize the decoder's image reconstruction error using a binary cross entropy loss. Secondly, we add an entropy loss term with weighting $\alpha$ that increases over time.
Its schedule is based on the current training epoch $e_c$  and the terminal training epoch $e_T$ as follows
\begin{equation}
	\alpha = 
	\begin{cases}
		0 & \text{if } e_c < 0.9\cdot e_T, \\
		5 \times 10^{-6} & \text{otherwise},
	\end{cases}
\end{equation}
such that the entropy of our latent representation decreases at the end of training to a minimum. This is necessary, since each observation should be precisely projected to only one latent sample in order to use return strategies from Go-Explore \citep{ecoffet2021first} but also to build directed graphs without multiple nodes for a single observation.  

\textbf{Dynamics Model.} The backwards world model uses a Multi-Layer Perceptron (MLP) \citep{rosenblatt1958perceptron} with SiLU nonlinearity \citep{elfwing2018sigmoid}  and layer normalization \citep{ba2016layer}. It takes an observation and action encoding $(z_t,a_{t-1})$ as input and outputs the estimated distribution $\hat{Z}_{t-1}$ over previous latent states. Consequently, the world model's loss minimizes the discrete Kullback–Leibler ($\mathrm{D_{KL}}$) divergence


\begin{equation}
	\begin{aligned}
		\mathcal{L}^{\text{WM}}_\theta =  \: \mathbb{E}_t\Bigl[ &\mathrm{D_{KL}}\bigl(\hat{Z}_{t-1} \| Z_{t-1}\bigr)\Bigr], \\
		& \text{with}\:\: \Phi_\theta(s_{t-1}) = Z_{t-1}, \\
		& \text{and}\:\: \Psi_\theta(z_t,a_{t-1}) = \hat{Z}_{t-1}
	\end{aligned}
\end{equation}

between the encoder's latent distribution $ Z_{t-1} $ and the world model's predicted distribution $ \hat{Z}_{t-1} $. Both models are jointly trained to allow  gradient flow back into the encoder parameters and therefore allowing the world model to influence the encoding structure in the latent space. The complete loss function is written as

\begin{equation}
	\mathcal{L}^{\text{}}_\theta = \mathcal{L}^{\text{Repr}}_\theta + w_{\text{wm}} \cdot \mathcal{L}^{\text{WM}}_\theta,
\end{equation} 
 which combines both terms with an additional weight $w_{\text{wm}}=0.0025$ for the world model loss.

\textbf{Policy optimization.} With a \emph{discrete} world model at hand, we create an improved dataset $\mathcal{D}$ for imitation learning instead of just relying on raw rollout trajectories from a (non-discrete) world model. The following procedure is implemented:
\begin{enumerate}
    \item Generate backward simulations via random action selection while starting at the goal $z^*$. In addition, start further rollouts at already simulated states based on their inverse visit frequency \citep[following the cell selection rule from][]{ecoffet2021first}.
    \item Create a directed graph (DG) with the simulation data where each node is a state encoding $z_t$. A \emph{directed} edge is added to the graph when a backward state transition $(z_{t+1},z_{t})$ is found in the simulations. In that way, the DG contains and unifies the information from all separate simulations.
    \item Apply a shortest path finding algorithm on the DG to find the shortest distances to the goal node.  For our method we use Dijkstra's algorithm \citep{dijkstra1959note} to get a shortest path estimate (SPE) for every state in the graph, e.g., $\text{SPE}(z^*) = 0$.
\end{enumerate}
By using the SPE, we evaluate whether a state-action pair $(z_t,a_t)$ should be used for the imitation learning dataset $\mathcal{D}$: if a forward transition $(z_t,a_t,z_{t+1})$ brings us closer to the goal, we include it, i.e.
\begin{equation}
    (\text{SPE}(z_{t}) > \text{SPE}(z_{t+1})) \Rightarrow (z_t,a_t) \in \mathcal{D}. \label{SPE}
\end{equation}
Note, that this dataset $\mathcal{D}$ has no loops and contains only paths that directly lead to the goal (according to the world model). In fact, the rule selects only shortest paths since it uses Dijkstra's algorithm for evaluation.

Finally we use the standard imitation learning technique of maximizing the log-probability for the selected state-action pairs $(z_t,a_t)$, i.e., we minimize

\begin{equation}
	\mathcal{L}^{\text{policy}}_\theta =  \mathbb{E}_t\left[ - \log(\pi_\theta(a_t|z_t)) + c_1 S[\pi_\theta](z_t) \right] \:\: \text{with} \:\: (z_t,a_t) \in \mathcal{D},
\end{equation}

where we added a small action space entropy bonus $S$ \citep{schulman2017proximal} with weight $c_1$. The entropy bonus is helpful if $\mathcal{D}$ contains bad samples. Furthermore, it makes additional policy improvement easier, e.g., by applying further classical reinforcement learning using REINFORCE \citep{sutton2018reinforcement} on the policy $\pi_\theta$.

\section{Experiments}

\begin{figure}
    \centering
    \begin{subfigure}{0.3\linewidth}
        \includegraphics[width=\linewidth]{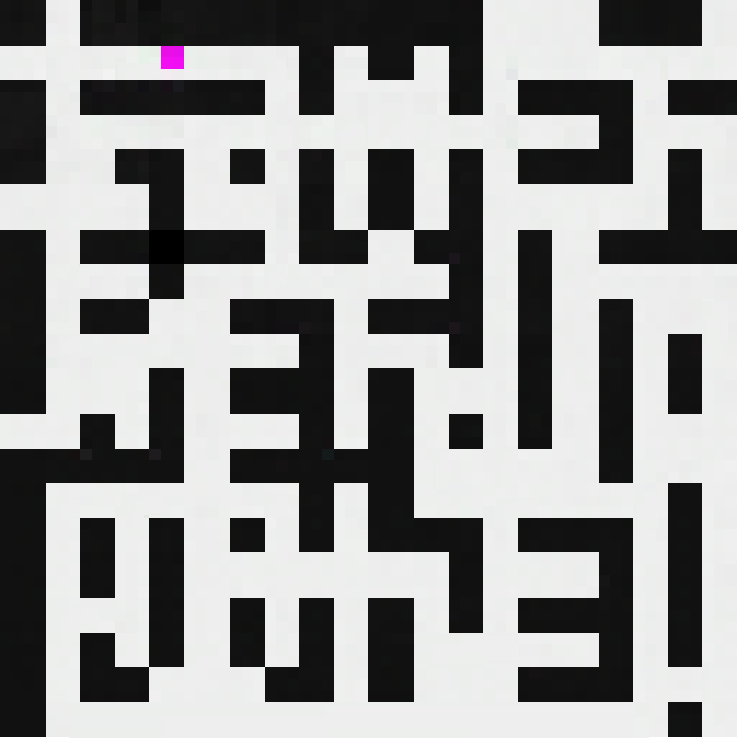}
        \caption{Maze (20$\times$20) observation}
        \label{fig:maze_obs}
    \end{subfigure}
    \hfill
    \begin{subfigure}{0.3\linewidth}
        \includegraphics[width=\linewidth]{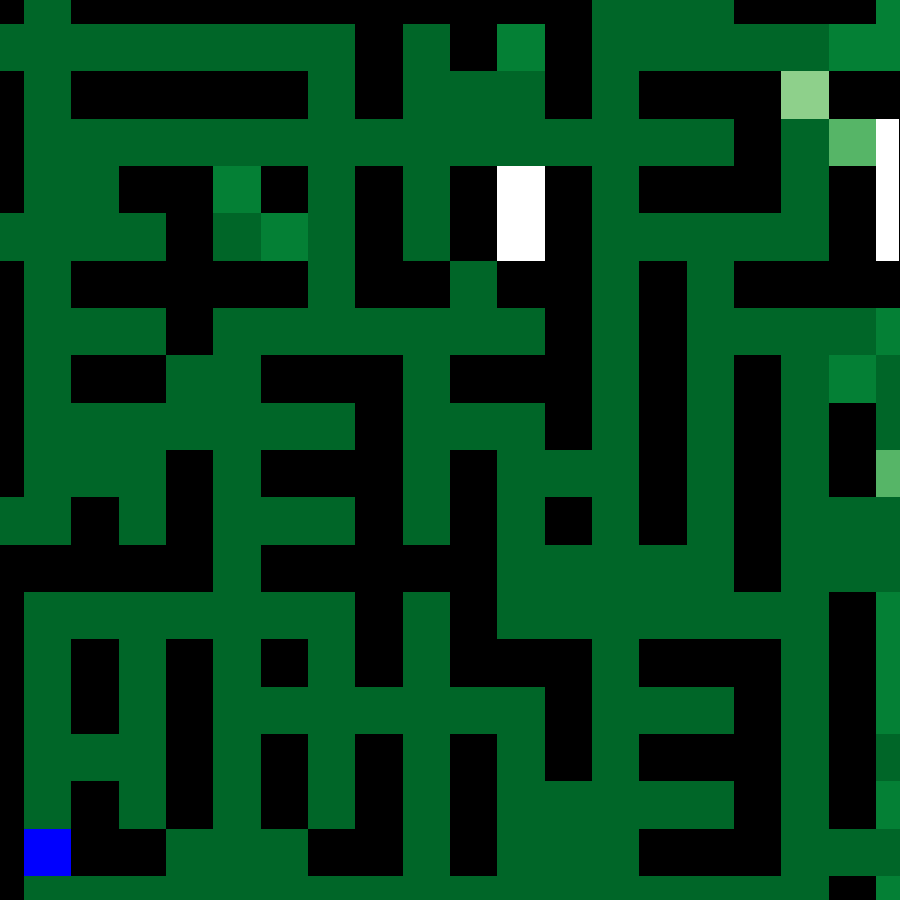}
        \caption{Single goal policy}
        \label{fig:single_goal}
    \end{subfigure}
    \hfill
    \begin{subfigure}{0.3\linewidth}
        \includegraphics[width=\linewidth]{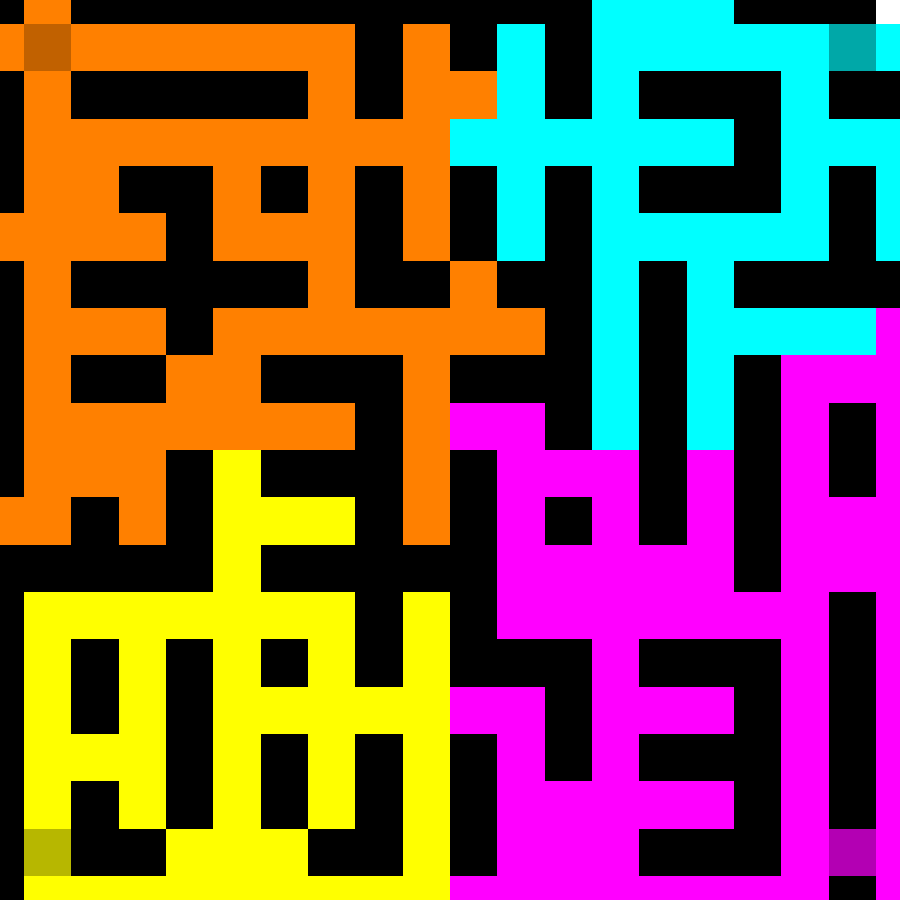}
        \caption{Four goal policy}
        \label{fig:multi_goal}
    \end{subfigure}
    \caption{Visualization of the maze environment. Panel (a) shows a raw observation where no goals are marked. Panel (b) shows the results for a learned policy with one goal state (marked as \textcolor{blue}{blue}).  The brightness of a square in \textcolor{greenfield}{green} indicates how often we reached the goal. Panel (c) shows results for four goal states (marked as dark-\{\textcolor{orangegoal}{orange},\textcolor{cyanfield}{cyan},\textcolor{yellowgoal}{yellow},\textcolor{violetgoal}{violet}\}). Colors indicate to which goal the policy has returned to, e.g., a position is colored \textcolor{orangefield}{orange} when the agent went to the \textcolor{orangegoal}{dark-orange} goal.}
    \label{fig:maze_eval}
\end{figure}

We study our method on a 20x20 maze environment where an observation is one $64 \times 64$ pixel image with RGB color channels and has a bird's-eye view over the whole maze (see Fig.~\ref{fig:maze_obs}). Note, that a deterministic maze is \emph{non-deterministic} when it is unrolled backwards in time. Our latent world model incorporates this property by predicting a latent state distribution \citep[as in][]{hafner2023mastering}.  For this work, the encoder, the decoder, the world model and the policy are neural networks.

After the world model training, every learned policy is conditioned on up to four goals. For our test, we choose the goal locations at the maze corners shown in
Fig.~1b/c. To deal with multiple goals for one policy requires only tiny adjustments in the DG. Finally, all policies are tested by placing the agent at every maze position five times and check whether they can return to their goal. To exclude attempts that reach the goal in time by chance, we consider only trials to be successful, if a goal was reached within 1.5 times the shortest goal distance. The locations marked in lighter green or even white indicate that the agent did not always reach the goal (see Fig.~1b/c).

Fig.~\ref{fig:quantative_result} shows the quantitative evaluation on the maze.  Policies are conditioned either on one goal (bottom left corner in Fig.~\ref{fig:single_goal}) or on four goals simultaneously (all corners in Fig.~\ref{fig:multi_goal}). The results show that the policies can reach their goal(s) from almost every maze position which are up to 37 steps away. Our experiments show we can learn complex goal-oriented policies even without rewards.

\section{Conclusion}
We introduce a novel method for goal-conditioning a policy on one or more states by using a discrete world model jointly with graph search.
Even though our method is not using rewards for optimization, it is still able to reach the desired state configurations by using a refined imitation learning dataset that has been generated by our backward world model.  

\begin{figure}
    \small
    \renewcommand{\arraystretch}{1.1}
    \centering
    \begin{subfigure}{0.42\linewidth}
        \centering
        \caption{One goal optimization}
        \begin{tabular}{|c|c|c|}
            \hline
            \setlength{\tabcolsep}{0pt}
            Maze size      & Goal                                     & Return positions   \\
            \hline
                           & \textcolor{orangegoal}{Orange}           & 207.1 / 234 (88\%) \\
            20 $\times$ 20 & \textcolor{cyanfield}{Cyan}              & 218.5 / 234 (93\%) \\
                           & \textcolor{yellowfield!70!black}{Yellow} & 210.2 / 234 (90\%) \\
                           & \textcolor{violetgoal}{Violet}           & 207.7 / 234 (89\%) \\
            \hline
        \end{tabular}
    \end{subfigure}
    \hfill
    \begin{subfigure}{0.57\linewidth}
        \centering
        \caption{Four goal optimization}
        \begin{tabular}{|c|c|c|}
            \hline
            \setlength{\tabcolsep}{0pt}
            Maze size      & Return positions   & Went to closest goal \\
            \hline
            20 $\times$ 20 & 221.9 / 234 (95\%) & 219.9 / 221.9 (99\%) \\
            \hline
        \end{tabular}
    \end{subfigure}
    \caption{Quantitative evaluation on the maze shown in Figure \ref{fig:maze_eval} with 20 runs for each result. One and four goal optimization use the goals from Figure \ref{fig:multi_goal}.}
    \label{fig:quantative_result}
\end{figure}

\bibliography{gcrl_workshop}
\bibliographystyle{gcrl_workshop}

\end{document}

%% file: math_commands.tex
\usepackage{amsmath,amsfonts,bm}

\def\eqref#1{equation~\ref{#1}}

\def\1{\bm{1}}

\DeclareMathAlphabet{\mathsfit}{\encodingdefault}{\sfdefault}{m}{sl}
\SetMathAlphabet{\mathsfit}{bold}{\encodingdefault}{\sfdefault}{bx}{n}













%% file: gcrl_workshop.bbl
\begin{thebibliography}{34}
\providecommand{\natexlab}[1]{#1}
\providecommand{\url}[1]{\texttt{#1}}
\expandafter\ifx\csname urlstyle\endcsname\relax
  \providecommand{\doi}[1]{doi: #1}\else
  \providecommand{\doi}{doi: \begingroup \urlstyle{rm}\Url}\fi

\bibitem[Andrychowicz et~al.(2017)Andrychowicz, Wolski, Ray, Schneider, Fong,
  Welinder, McGrew, Tobin, Pieter~Abbeel, and
  Zaremba]{andrychowicz2017hindsight}
Marcin Andrychowicz, Filip Wolski, Alex Ray, Jonas Schneider, Rachel Fong,
  Peter Welinder, Bob McGrew, Josh Tobin, OpenAI Pieter~Abbeel, and Wojciech
  Zaremba.
\newblock Hindsight experience replay.
\newblock \emph{Advances in neural information processing systems}, 30, 2017.

\bibitem[Ba et~al.(2016)Ba, Kiros, and Hinton]{ba2016layer}
Jimmy~Lei Ba, Jamie~Ryan Kiros, and Geoffrey~E Hinton.
\newblock Layer normalization.
\newblock \emph{arXiv preprint arXiv:1607.06450}, 2016.

\bibitem[Ballard(1987)]{ballard1987modular}
Dana~H Ballard.
\newblock Modular learning in neural networks.
\newblock In \emph{Proceedings of the sixth National Conference on artificial
  intelligence-volume 1}, pp.\  279--284, 1987.

\bibitem[Brown et~al.(2020)Brown, Mann, Ryder, Subbiah, Kaplan, Dhariwal,
  Neelakantan, Shyam, Sastry, Askell, et~al.]{brown2020language}
Tom Brown, Benjamin Mann, Nick Ryder, Melanie Subbiah, Jared~D Kaplan, Prafulla
  Dhariwal, Arvind Neelakantan, Pranav Shyam, Girish Sastry, Amanda Askell,
  et~al.
\newblock Language models are few-shot learners.
\newblock \emph{Advances in neural information processing systems},
  33:\penalty0 1877--1901, 2020.

\bibitem[Chebotar et~al.(2021)Chebotar, Hausman, Lu, Xiao, Kalashnikov, Varley,
  Irpan, Eysenbach, Julian, Finn, et~al.]{chebotar2021actionable}
Yevgen Chebotar, Karol Hausman, Yao Lu, Ted Xiao, Dmitry Kalashnikov, Jake
  Varley, Alex Irpan, Benjamin Eysenbach, Ryan Julian, Chelsea Finn, et~al.
\newblock Actionable models: Unsupervised offline reinforcement learning of
  robotic skills.
\newblock \emph{arXiv preprint arXiv:2104.07749}, 2021.

\bibitem[Chelu et~al.(2020)Chelu, Precup, and van
  Hasselt]{chelu2020forethought}
Veronica Chelu, Doina Precup, and Hado~P van Hasselt.
\newblock Forethought and hindsight in credit assignment.
\newblock \emph{Advances in Neural Information Processing Systems},
  33:\penalty0 2270--2281, 2020.

\bibitem[Dijkstra(1959)]{dijkstra1959note}
Edsger~W Dijkstra.
\newblock A note on two problems in connexion with graphs.
\newblock \emph{Numerische mathematik}, 1\penalty0 (1):\penalty0 269--271,
  1959.

\bibitem[Ecoffet et~al.(2021)Ecoffet, Huizinga, Lehman, Stanley, and
  Clune]{ecoffet2021first}
Adrien Ecoffet, Joost Huizinga, Joel Lehman, Kenneth~O Stanley, and Jeff Clune.
\newblock First return, then explore.
\newblock \emph{Nature}, 590\penalty0 (7847):\penalty0 580--586, 2021.

\bibitem[Edwards et~al.(2018)Edwards, Downs, and Davidson]{edwards2018forward}
Ashley~D Edwards, Laura Downs, and James~C Davidson.
\newblock Forward-backward reinforcement learning.
\newblock \emph{arXiv preprint arXiv:1803.10227}, 2018.

\bibitem[Elfwing et~al.(2018)Elfwing, Uchibe, and Doya]{elfwing2018sigmoid}
Stefan Elfwing, Eiji Uchibe, and Kenji Doya.
\newblock Sigmoid-weighted linear units for neural network function
  approximation in reinforcement learning.
\newblock \emph{Neural networks}, 107:\penalty0 3--11, 2018.

\bibitem[Goyal et~al.(2018)Goyal, Brakel, Fedus, Singhal, Lillicrap, Levine,
  Larochelle, and Bengio]{goyal2018recall}
Anirudh Goyal, Philemon Brakel, William Fedus, Soumye Singhal, Timothy
  Lillicrap, Sergey Levine, Hugo Larochelle, and Yoshua Bengio.
\newblock Recall traces: Backtracking models for efficient reinforcement
  learning.
\newblock \emph{arXiv preprint arXiv:1804.00379}, 2018.

\bibitem[Gupta et~al.(2019)Gupta, Kumar, Lynch, Levine, and
  Hausman]{gupta2019relay}
Abhishek Gupta, Vikash Kumar, Corey Lynch, Sergey Levine, and Karol Hausman.
\newblock Relay policy learning: Solving long-horizon tasks via imitation and
  reinforcement learning.
\newblock \emph{arXiv preprint arXiv:1910.11956}, 2019.

\bibitem[Ha \& Schmidhuber(2018)Ha and Schmidhuber]{ha2018world}
David Ha and J{\"u}rgen Schmidhuber.
\newblock World models.
\newblock \emph{arXiv preprint arXiv:1803.10122}, 2018.

\bibitem[Hafner et~al.(2020)Hafner, Lillicrap, Norouzi, and
  Ba]{hafner2020mastering}
Danijar Hafner, Timothy Lillicrap, Mohammad Norouzi, and Jimmy Ba.
\newblock Mastering atari with discrete world models.
\newblock \emph{arXiv preprint arXiv:2010.02193}, 2020.

\bibitem[Hafner et~al.(2022)Hafner, Lee, Fischer, and Abbeel]{hafner2022deep}
Danijar Hafner, Kuang-Huei Lee, Ian Fischer, and Pieter Abbeel.
\newblock Deep hierarchical planning from pixels.
\newblock \emph{Advances in Neural Information Processing Systems},
  35:\penalty0 26091--26104, 2022.

\bibitem[Hafner et~al.(2023)Hafner, Pasukonis, Ba, and
  Lillicrap]{hafner2023mastering}
Danijar Hafner, Jurgis Pasukonis, Jimmy Ba, and Timothy Lillicrap.
\newblock Mastering diverse domains through world models.
\newblock \emph{arXiv preprint arXiv:2301.04104}, 2023.

\bibitem[H{\"o}ftmann et~al.(2023)H{\"o}ftmann, Robine, and
  Harmeling]{hoftmann2023time}
Marc H{\"o}ftmann, Jan Robine, and Stefan Harmeling.
\newblock Time-myopic go-explore: Learning a state representation for the
  go-explore paradigm.
\newblock \emph{arXiv preprint arXiv:2301.05635}, 2023.

\bibitem[Huang \& Rao(2011)Huang and Rao]{huang2011predictive}
Yanping Huang and Rajesh~PN Rao.
\newblock Predictive coding.
\newblock \emph{Wiley Interdisciplinary Reviews: Cognitive Science}, 2\penalty0
  (5):\penalty0 580--593, 2011.

\bibitem[Kaiser et~al.(2019)Kaiser, Babaeizadeh, Milos, Osinski, Campbell,
  Czechowski, Erhan, Finn, Kozakowski, Levine, et~al.]{kaiser2019model}
Lukasz Kaiser, Mohammad Babaeizadeh, Piotr Milos, Blazej Osinski, Roy~H
  Campbell, Konrad Czechowski, Dumitru Erhan, Chelsea Finn, Piotr Kozakowski,
  Sergey Levine, et~al.
\newblock Model-based reinforcement learning for atari.
\newblock \emph{arXiv preprint arXiv:1903.00374}, 2019.

\bibitem[Kidambi et~al.(2020)Kidambi, Rajeswaran, Netrapalli, and
  Joachims]{kidambi2020morel}
Rahul Kidambi, Aravind Rajeswaran, Praneeth Netrapalli, and Thorsten Joachims.
\newblock Morel: Model-based offline reinforcement learning.
\newblock \emph{Advances in neural information processing systems},
  33:\penalty0 21810--21823, 2020.

\bibitem[Lai et~al.(2020)Lai, Shen, Zhang, and Yu]{lai2020bidirectional}
Hang Lai, Jian Shen, Weinan Zhang, and Yong Yu.
\newblock Bidirectional model-based policy optimization.
\newblock In \emph{International Conference on Machine Learning}, pp.\
  5618--5627. PMLR, 2020.

\bibitem[LeCun et~al.(1989)LeCun, Boser, Denker, Henderson, Howard, Hubbard,
  and Jackel]{lecun1989backpropagation}
Yann LeCun, Bernhard Boser, John~S Denker, Donnie Henderson, Richard~E Howard,
  Wayne Hubbard, and Lawrence~D Jackel.
\newblock Backpropagation applied to handwritten zip code recognition.
\newblock \emph{Neural computation}, 1\penalty0 (4):\penalty0 541--551, 1989.

\bibitem[Li et~al.(2022)Li, Tang, Tomizuka, and Zhan]{li2022hierarchical}
Jinning Li, Chen Tang, Masayoshi Tomizuka, and Wei Zhan.
\newblock Hierarchical planning through goal-conditioned offline reinforcement
  learning.
\newblock \emph{IEEE Robotics and Automation Letters}, 7\penalty0 (4):\penalty0
  10216--10223, 2022.

\bibitem[Pan \& Lin(2022)Pan and Lin]{pan2022backward}
Yuxin Pan and Fangzhen Lin.
\newblock Backward imitation and forward reinforcement learning via
  bi-directional model rollouts.
\newblock In \emph{2022 IEEE/RSJ International Conference on Intelligent Robots
  and Systems (IROS)}, pp.\  9040--9047. IEEE, 2022.

\bibitem[Pateria et~al.(2021)Pateria, Subagdja, Tan, and
  Quek]{pateria2021hierarchical}
Shubham Pateria, Budhitama Subagdja, Ah-hwee Tan, and Chai Quek.
\newblock Hierarchical reinforcement learning: A comprehensive survey.
\newblock \emph{ACM Computing Surveys (CSUR)}, 54\penalty0 (5):\penalty0 1--35,
  2021.

\bibitem[Robine et~al.(2023)Robine, H{\"o}ftmann, Uelwer, and
  Harmeling]{robine2023transformer}
Jan Robine, Marc H{\"o}ftmann, Tobias Uelwer, and Stefan Harmeling.
\newblock Transformer-based world models are happy with 100k interactions.
\newblock \emph{arXiv preprint arXiv:2303.07109}, 2023.

\bibitem[Rosenblatt(1958)]{rosenblatt1958perceptron}
Frank Rosenblatt.
\newblock The perceptron: a probabilistic model for information storage and
  organization in the brain.
\newblock \emph{Psychological review}, 65\penalty0 (6):\penalty0 386, 1958.

\bibitem[Schmidhuber(1991)]{schmidhuber1991possibility}
J{\"u}rgen Schmidhuber.
\newblock A possibility for implementing curiosity and boredom in
  model-building neural controllers.
\newblock In \emph{Proc. of the international conference on simulation of
  adaptive behavior: From animals to animats}, pp.\  222--227, 1991.

\bibitem[Schrittwieser et~al.(2020)Schrittwieser, Antonoglou, Hubert, Simonyan,
  Sifre, Schmitt, Guez, Lockhart, Hassabis, Graepel,
  et~al.]{schrittwieser2020mastering}
Julian Schrittwieser, Ioannis Antonoglou, Thomas Hubert, Karen Simonyan,
  Laurent Sifre, Simon Schmitt, Arthur Guez, Edward Lockhart, Demis Hassabis,
  Thore Graepel, et~al.
\newblock Mastering atari, go, chess and shogi by planning with a learned
  model.
\newblock \emph{Nature}, 588\penalty0 (7839):\penalty0 604--609, 2020.

\bibitem[Schulman et~al.(2017)Schulman, Wolski, Dhariwal, Radford, and
  Klimov]{schulman2017proximal}
John Schulman, Filip Wolski, Prafulla Dhariwal, Alec Radford, and Oleg Klimov.
\newblock Proximal policy optimization algorithms.
\newblock \emph{arXiv preprint arXiv:1707.06347}, 2017.

\bibitem[Sutton \& Barto(2018)Sutton and Barto]{sutton2018reinforcement}
Richard~S Sutton and Andrew~G Barto.
\newblock \emph{Reinforcement learning: An introduction}.
\newblock MIT press, 2018.

\bibitem[Wang et~al.(2021{\natexlab{a}})Wang, Li, Jiang, Zhu, Li, and
  Zhang]{wang2021offline}
Jianhao Wang, Wenzhe Li, Haozhe Jiang, Guangxiang Zhu, Siyuan Li, and Chongjie
  Zhang.
\newblock Offline reinforcement learning with reverse model-based imagination.
\newblock \emph{Advances in Neural Information Processing Systems},
  34:\penalty0 29420--29432, 2021{\natexlab{a}}.

\bibitem[Wang et~al.(2021{\natexlab{b}})Wang, Xu, Du, and
  Lee]{wang2021learning}
Yunke Wang, Chang Xu, Bo~Du, and Honglak Lee.
\newblock Learning to weight imperfect demonstrations.
\newblock In \emph{International Conference on Machine Learning}, pp.\
  10961--10970. PMLR, 2021{\natexlab{b}}.

\bibitem[Wu et~al.(2019)Wu, Charoenphakdee, Bao, Tangkaratt, and
  Sugiyama]{wu2019imitation}
Yueh-Hua Wu, Nontawat Charoenphakdee, Han Bao, Voot Tangkaratt, and Masashi
  Sugiyama.
\newblock Imitation learning from imperfect demonstration.
\newblock In \emph{International Conference on Machine Learning}, pp.\
  6818--6827. PMLR, 2019.

\end{thebibliography}
